\documentclass[journal,twoside]{IEEEtran}
\usepackage{amsmath,amssymb,amsfonts}
\usepackage{algorithm}

\usepackage{algpseudocode}
\usepackage{graphicx}
\usepackage{textcomp}
\usepackage{listings}
\usepackage{booktabs}
\usepackage{comment}
\usepackage{cleveref}
\newcommand{\ie}{i.e.\ }

\makeatletter
\def\ps@IEEEtitlepagestyle{%
  \def\@oddhead{\normalfont\scriptsize \hfill\thepage}%
  \def\@evenhead{\normalfont\scriptsize \hfill\thepage}%
  \def\@oddfoot{}\def\@evenfoot{}}
\makeatother

\begin{document}
\title{Uncertainty Estimation for Trust Attribution \\ to Speed-of-Sound Reconstruction with \\ Variational Networks}

\author{Sonia Laguna, Lin Zhang, Can Deniz Bezek, Monika Farkas, Dieter Schweizer, \\Rahel A. Kubik-Huch,  and Orcun Goksel
\thanks{Sonia Laguna, Lin Zhang, Dieter Schweizer, and Orcun Goksel are with the Computer-assisted Applications in Medicine, ETH Zurich, Switzerland.}
\thanks{Can Deniz Bezek and Orcun Goksel are with the Department of Information Technology, Uppsala University, Sweden.}
\thanks{Monika Farkas and Rahel A. Kubik-Huch are with the Department of Radiology, Kantonsspital Baden, affiliated Hospital for Research and Teaching of the Faculty of Medicine of the University of Zurich, Switzerland.}
\thanks{Computation hardware was supported by a grant from Nvidia and funding was provided partially by Uppsala Medtech Science \& Innovation Centre.}\vspace*{-3.5ex}
}
\maketitle

\begin{abstract}

Speed-of-sound (SoS) is a biomechanical characteristic of tissue, and its imaging can provide a promising biomarker for diagnosis. 
Reconstructing SoS images from ultrasound acquisitions can be cast as a limited-angle computed-tomography problem, with Variational Networks being a promising model-based deep learning solution.  
Some acquired data frames may, however, get corrupted by noise due to, e.g., motion, lack of contact, and acoustic shadows, which in turn negatively affects the resulting SoS reconstructions. 
We propose to use the uncertainty in SoS reconstructions to attribute trust to each individual acquired frame. 
Given multiple acquisitions, we then use an uncertainty based automatic selection among these retrospectively, to improve diagnostic decisions. 
We investigate uncertainty estimation based on Monte Carlo Dropout and Bayesian Variational Inference. We assess our automatic frame selection method for differential diagnosis of breast cancer, distinguishing between benign fibroadenoma and malignant carcinoma. 
We evaluate 21 lesions classified as BI-RADS~4, which represents suspicious cases for probable malignancy. The most trustworthy frame among four acquisitions of each lesion was identified using uncertainty based criteria. 
Selecting a frame informed by uncertainty achieved an area under curve of 76\% and 80\% for Monte Carlo Dropout and Bayesian Variational Inference, respectively, superior to any uncertainty-uninformed baselines with the best one achieving 64\%. 
A novel use of uncertainty estimation is proposed for selecting one of multiple data acquisitions for further processing and decision making.
\end{abstract}

\begin{IEEEkeywords}
Ultrasonography, image reconstruction, breast cancer differential diagnosis.
\end{IEEEkeywords}

\section{Introduction}
\label{sec:introduction}

Breast cancer is a leading cause of cancer-related mortality in women~\cite{houghton2021cancer}.
Ductal Carcinoma (CA) is the most frequent type of malignant breast lesion, 
while Fibroadenoma (FA) is the most common type of benign breast lesion~\cite{guray2006benign}. 
In breast cancer, early detection is the key in reducing mortality rate~\cite{houghton2021cancer}, emphasizing the need for fast, reliable, and efficient techniques to detect and classify lesions.
Current gold-standard diagnostic and screening methods, such as mammography and MRI, have several limitations, e.g., ionizing-radiation, not being in real-time, and reduced sensitivity with increased breast density or increased parencyhmal background enhancement~\cite{bae2014breast}. 
While biopsy can provide exact tissue information, it is localized only in several tissue points, and carries a risk of complications and places a burden on both the patient and the clinical pathology pipelines.

Ultrasound (US) offers a non-ionizing, real-time, and cost-effective imaging alternative, but conventional B-mode images do not provide sufficient information for differentiating malignant lesions. 
Speed-of-sound (SoS) is a promising quantitative biomechanical marker that can provide information about the pathological state of the tissue.
Reconstructing the SoS distribution was proposed using through-transmission US computed tomography.  
However, such systems necessitate a complex hardware setup, submersion of the breast in water, and a trained operator -- complicating the clinical routine.
Pulse-echo methods eliminate the aforementioned limitations and enable SoS estimation with conventional ultrasound transducers. 
Some of these use echo profiles~\cite{anderson_direct_1998}, optimize an image quality metric~\cite{perrot_so_2021}, relate the speckle-shifts between different transmissions to SoS~\cite{jaeger2015computed,sanabria2018spatial,rau2021speed,schweizer2023robust, bezek2024model, bezek_breast_2025}, or reconstruct SoS with black-box networks~\cite{feigin2019deep}.
A conventional hand-held US transducer operated in pulse-echo mode has shown promising results in differentiating CA and FA, using the distinct SoS contrasts of these inclusions compared to background~\cite{ruby2019breast}.

US imaging with pulse-echo SoS is an ill-posed limited angle inverse problem, which is typically solved analytically with hand-crafted regularizers~\cite{rau2021speed,schweizer2023robust}.
Black-box deep learning~\cite{zhu2018image, feigin2019deep} or pre-/post-processing of analytical solutions with neural networks~\cite{jin2017deep} have been proposed for image reconstruction.
Yet, in cases where annotated \emph{in vivo} data are not available, model-based deep learning through loop unrolling was found to be more effective~\cite{mccann2019biomedical}. 
Variational Network (VN) is a model-based image reconstruction framework and enables prior learning in restricted analytical forms within algorithm loop unrolling~\cite{monga2021algorithm}.
This allows generalizability to \emph{in vivo} from training from only simulated data, with its success shown for various imaging inverse problems, e.g., compressed sensing for MRI~\cite{hammernik_learning_2018}, limited-angle X-ray computed tomography~\cite{vishnevskiy2019deep}, and imaging SoS~\cite{bernhardt2020training}. 
As the output of a deep learning method may vary largely depending on various aspects, estimating uncertainty in such outputs is of utmost importance, especially for high-risk applications such as medical imaging and decision making, as was studied for MRI~\cite{schlemper2018bayesian,narnhofer2021bayesian,ekmekci2022uncertainty}.

\begin{figure*}
\centerline{\includegraphics[width=\textwidth]{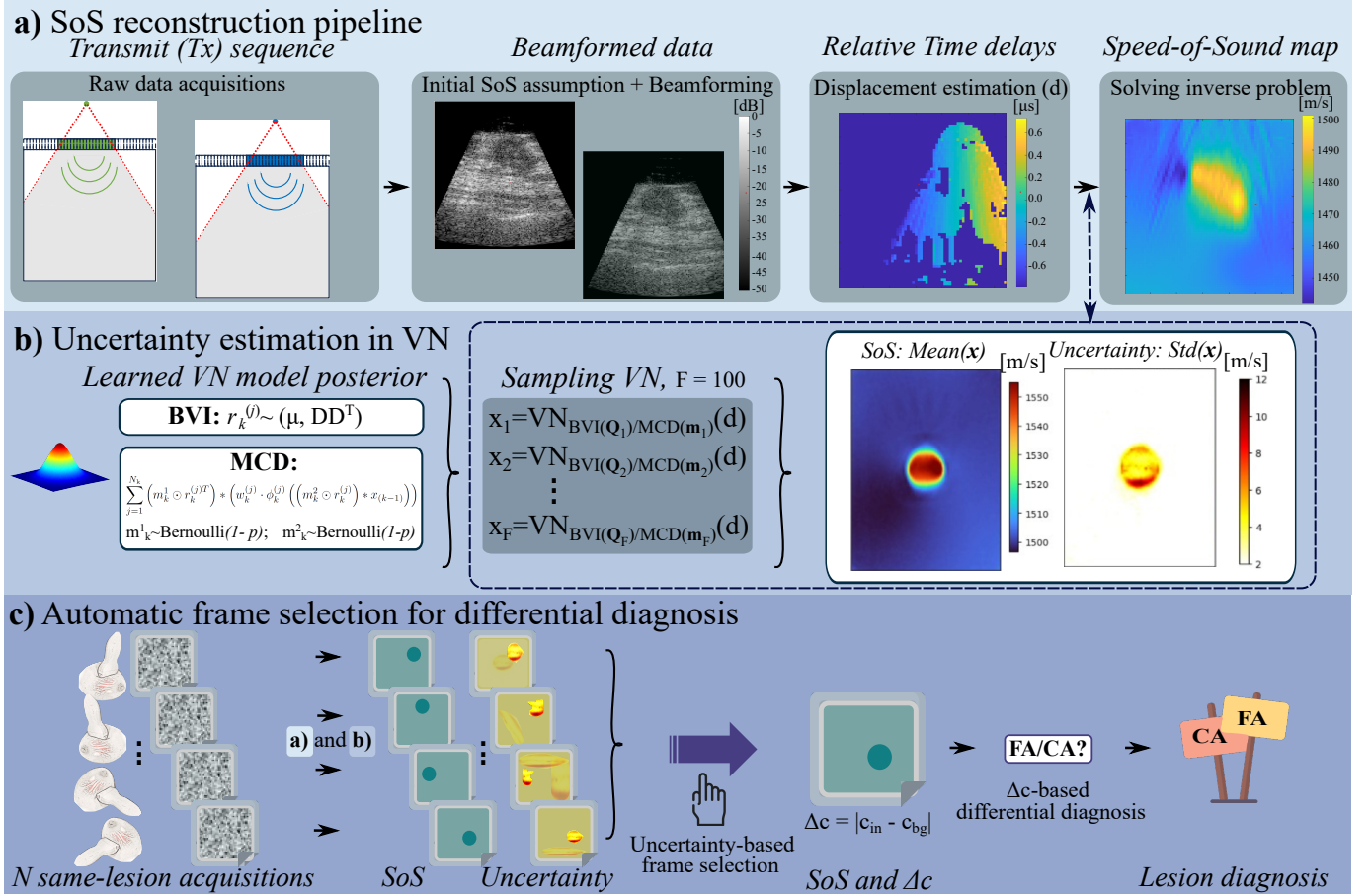}}
\caption{a) SoS reconstruction pipeline on a clinical example: Echo data from different transmit events are beamformed, between pairs of which the displacements are estimated. 
Using multiple displacement maps, a SoS map is reconstructed by solving an inverse problem with VN.
b) The VN posterior is learned using Bayesian variational inference or Monte Carlo Dropout.
At inference, samples are drawn from each posterior, with their mean being the reconstructed SoS image and the standard deviation the uncertainty estimate. 
c) Data from the same lesion is collected multiple times. 
Reconstruction uncertainty is used to select an optimal acquisition, later used for breast cancer differential diagnosis.}
\label{fig:pipeline_sos}
\end{figure*}

In this work, we study uncertainty estimation in SoS imaging, with the following key contributions: (1) we propose a novel use of uncertainty estimates as trust attribution to select the best acquisition frame in US imaging, leveraging its rapid and interactive acquisition capability; (2) we apply uncertainty estimation to SoS reconstruction for the first time; (3) we introduce a “relative” (normalized) uncertainty metric superior in trust attribution in regression-like tasks; (4) we demonstrate the clinical feasibility of this approach for breast cancer diagnosis with SoS imaging, including the first in-vivo application of VN for SoS reconstruction; and (5) we present an algebraic reformulation for efficient uncertainty computation for large models on standard hardware. We implement this based on VN models and using well-studied uncertainty estimation frameworks, Monte Carlo Dropout and Bayesian Variational Inference~\cite{narnhofer2021bayesian,gal2016dropout}. These innovations advance the reliability and clinical applicability of SoS. The framework is illustrated in \Cref{fig:pipeline_sos}.
\section{Methods} 
\subsection{SoS Imaging and Variational Network}
\label{SoS_Imaging_Princ}

The approach in this work relies on measuring speckle shifts between beamformed frames obtained from different transmit/receive (Tx/Rx) sequences. 
A speckle observed under varying Tx/Rx parameters appears at shifted locations if the SoS changes along the propagation paths associated with these parameters. 
Our method relates local SoS variations, inferred from the beamforming SoS, to the observed echo shifts along different propagation paths, using an imaging model that accounts for the wave propagation. 
To reconstruct SoS distributions, beamformed images are generated using distinct Tx/Rx parameters, which are selected to ensure that the resulting echo shifts are sensitive to spatial SoS variations. Images are then beamformed from the acquired raw data using an assumed tissue SoS value ($c_0$). 
Apparent speckle shifts (displacements) are found between pairs of $M$ beamformed frames (see \Cref{fig:pipeline_sos} (a)). 
Finally, a SoS map is reconstructed by solving the following inverse problem:
$
    {{\hat{x}}} = \arg\min_{{{x}}} \|L({{x} - x_{\mathrm{0}})} - {{d}}\|_p + \lambda \mathcal{R}({x}),\ 
$
where ${x} \in \mathbb{R}^{N_\text{x}N_\text{z}}$ is the sought (vectorized) slowness (inverse SoS) map on a discretized $N_\text{x} \times N_\text{z}$ reconstruction grid, and $x_{\mathrm{0}}$ is the initial beamforming slowness ($1 / c_{\mathrm{0}})$.
Vectorized time delay (displacement) measurements on a $N_\text{x'} \times N_\text{z'}$ grid concatenated from $M$ image pairs are denoted by ${d} \in \mathbb{R}^{MN_\text{x'}N_\text{z'}}$. 
Sensitivity of measurements to SoS map is then encoded by a sparse forward-model matrix $L \in \mathbb{R}^{MN_\text{x'}N_\text{z'} \times  N_\text{x}N_\text{z}}$.
The first term in the inverse problem corresponds to data fidelity, for which a norm $p$ is chosen based on expected measurement noise, e.g., $l_2$-norm for Gaussian noise.
The second term corresponds to regularization, with $\lambda$ controlling its impact.
The above optimization is typically solved with iterative analytical methods, with LBFGS~\cite{broyden1970convergence} being an efficient option.
The SoS distribution is computed as $\hat{c} = 1/(\hat{x}+x_0)$. 

Conventional regularization priors penalize total variation of image gradients, e.g., via $l_1$-norm, i.e., $\lambda \| {R_\text{grad}} {{x}}\|_1$. 
A more general form can use multiple priors as field of experts~\cite{roth2009fields}, i.e., $\sum_{i=1}^N \lambda_i \| {R_i} {{x}}\|_{t_i}$.
For instance, total variation can be expressed as $\lambda \| r_\mathrm{x} \ast {{x}}\|_1 + \lambda \| r_\mathrm{y} \ast {{x}}\|_1$ based on lateral and axial gradient filters $r_\mathrm{x}$ and $r_\mathrm{y}$.
A useful subset of such linear priors can be effectively represented using convolutions, i.e., as \mbox{$\lambda_i \| r_i \ast {{x}}\|_{t_i}$} with $\ast$ denoting convolution and $r_i$ the kernels. 
This form then can be extended with spatially-varying weighting (instead of a fixed $\lambda$ for the entire image) and arbitrary penalty transformations (instead of hand-crafted norms). 
This can then model a wide range of useful regularizers in the form of $w_i\cdot\phi_i\text{(}r_i \ast {x}\text{)}$, where $w_i\in\mathbb{R}^{a_1 a_2}$ is a vector with weights for each image location and 
$\phi_i$ is a transformation function.
To enable learning, $\phi_i$ for mapping (scaling) values is parameterized in a normalized space via some control points. 
This can then represent typical norms as well as non-norm functionals.

With its accuracy shown in various imaging inverse problems~\cite{bezek2024model,kobler2017variational,hammernik_learning_2018,vishnevskiy2019deep,bernhardt2020training}, VN is a model-based deep learning framework that learns data fitness and regularization in a restricted analytical form, via unrolling $K$ iterations of an optimization algorithm, e.g., the gradient descent for SoS as follows:
\begin{equation*}
    {x}_{k} = {x}_{(k-1)} - \Big[({s}_{k}{L})^\text{T}\psi_{k}({s}_{k}({L}{x}_{(k-1)} - {d})) + \lambda\nabla R({x}_{(k-1)}) \Big].
\end{equation*}
The first term corresponding to data gradient involves the penalty transformation $\psi(.)$ and the spatial weight $s_k$ (desired fidelity to each time delay measurement location) as learnable parameters at the $k$-th unrolled layer. 
The second term is the regularization gradient, in a restricted analytical form as: 
\begin{equation}
\nabla R(x_{k}) = \sum_{j=1}^{N_\text{k}} {r}_k^{(j) T}*\left({w}_k^{(j)}\cdot\phi_k^{(j)}\left({r}_k^{(j)}
*{x}_{(k-1)}\right)\right),
\label{reg_param}
\end{equation}
where $N_\text{k}$ is the total number of priors in the VN field of experts model.
A weighted exponential loss was applied across the unrolling layers~\cite{vishnevskiy2019deep}, see Appendix~\ref{app2}.

\subsection{Uncertainty estimation}
\label{sec:unc}

\subsubsection{Monte Carlo Dropout (MCD)}
Dropout is a technique that randomly deactivates a fraction of weights in a neural network with a specified probability, namely dropout rate.
It is commonly used during training to reduce overfitting.
Monte Carlo Dropout (MCD) applies dropout also at inference time, to sample multiple outcome hypotheses and construct the posterior distribution of a model~\cite{gal2016dropout}. 
Generally, black-box learning methods apply dropout at the individual weight level.
For the restricted analytical form of VN, however, dropping individual (e.g., kernel, spatial, transformation function control points) weights are not meaningful, since these are only relevant with respect to each other, e.g., within a kernel.
Accordingly, although our MCD is based on kernel (filter) removal, it corresponds to removing the respective prior component.
Analytically, the probability of a component remaining active in the VN follows a Bernoulli distribution with parameter \mbox{\textit{1 - p}}.
Concretely, we choose to dropout the filters, ${r}_k^{(j)}$, which are used to model the regularization term.
The filters are multiplied element-wise with the independent Bernoulli variables that act as a mask to the input variable:
\begin{equation*}
\label{drop_eq}
\sum_{j=1}^{N_\text{k}}\left( {m}_k^{1} \odot {r}_k^{(j) T} \right)\ast\left({w}_k^{(j)}\cdot\phi_k^{(j)}\left(\left({m}_k^{2} \odot {r}_k^{(j)}\right) \ast {x}_{(k-1)}\right)\right),
\end{equation*}
where $\mathrm{m}_k^{1}\sim\text {Bernoulli}(\mathit{1-p})$ and $\mathrm{m}_k^{2}\sim\text {Bernoulli}(\mathit{1-p})$ are the layer dropout masks, with each element being dropped with probability $\mathit{p}$.
To compute the final uncertainty and SoS reconstruction results, the models were sampled at test time.

\subsubsection{Bayesian Variational Inference (BVI)}

Bayesian Variational Inference (BVI) aims to model the posterior distribution of intrinsic model parameters. 
In particular, we learn the distribution of filters ${r}_k^{(j)}$.
Following~\cite{narnhofer2021bayesian}, we model the filters as a multivariate Gaussian distribution $N(\mu, \Sigma)$, with learned mean $\mu \in \mathbb{R}^t$ and learned covariance matrix $\Sigma \in \mathbb{R}^{t \times t}$, where t is the total number of all filters.
Since $\Sigma$ is a positive definite and symmetric matrix,
for computation and memory efficiency, we decompose it as $\Sigma=D D^{\top} \in \mathbb{R}^{t \times t}$, where $D \in \mathbb{R}^{t \times t}$ is the lower triangular Cholesky factor.
With the assumption of filters being uncorrelated with each other, $\Sigma$ admits a block diagonal structure, with $o^2$$\times$$o^2$ blocks, where $o$ is the filter size.
Samples from  ${r}_k^{(j)}$ can be directly obtained as $\mu+D y$, with $y \sim \mathcal{N}(0, \mathrm{I_{d}})$.

An additional Kullback–Leibler (KL) loss is used to prevent degenerate models. Particularly, it measures the difference between two Gaussian distributions $q_1$ and $q_2$: 
\begin{align}
\mathrm{KL}(q_1 \| q_2) &= \frac{1}{2}\log \frac{|\Sigma_2|}{\left|\Sigma_1\right|} + \operatorname{tr}(\Sigma_2^{-1} \Sigma_1)  \notag \\
&\quad +\left(\mu_2-\mu_1\right)^{\top} \Sigma_2^{-1}((\mu_2-\mu_1)-t) ,
\end{align}
where our sought distribution $q_1=\mathcal{N}(\mu, \Sigma)$ is regularized by normal distribution $q_2=\mathcal{N}\left(\mu, \alpha^{-1} \mathrm{I_{d}}\right)$, 
with $\alpha > 0$  being a hyperparameter as a prior.
Neglecting constant terms for optimization, $\mathrm{KL}(q_1 \| q_2)\approx\alpha \operatorname{tr}(D D^{\top})-\log(\operatorname{det}(D D^{\top}))$.
As the determinant of $DD^T$ is computationally expensive and intractable, in this paper we propose the following novel reformulation:
\begin{equation}
    \begin{aligned}
        \log \left(\operatorname{det}\left(D D^T\right)\right) 
        &= \log \left(\operatorname{det}(D)^2\right) \\
        &= 2 \log \prod_i D_{i i} \\
        &= 2 \sum_i \log \left(D_{i i}\right) \\
        &= 2 \operatorname{tr}(\log (D)).
    \end{aligned}
    \label{log_det_eq}
\end{equation}
Therefore,
$\mathrm{KL}(q_1 \| q_2)\approx\alpha \operatorname{tr}(D D^{\top})-2 \operatorname{tr}(\log (D))$,
and the final BVI loss becomes
\begin{align}
\mathcal{L}_\text{BVI} &= \sum_{k=1}^K \exp^{-\tau(K-k)}\left\|{x}_k-{x}^*\right\|_1  \notag \\
&\quad + \lambda_r \sum_{k=1}^K\sum_{j=1}^{N_\text{j}^{\phi_\text{k}}} \sqrt{\left(y_{j-1}^{\phi_k}-2 y_j^{\phi_k}+y_{j+1}^{\phi_k}\right)^2+\varepsilon} \notag \\
&\quad + \beta\left(\alpha \operatorname{tr}(D D^{\top})-2 \operatorname{tr}(\log (D))\right),
\end{align}
where $\beta > 0$ weights the KL term to balance its effect in the final loss.

MCD and BVI were sampled at test time to estimate (1)~the SoS reconstruction with their mean and (2)~uncertainty with their standard deviation (see \Cref{fig:pipeline_sos}\textbf{b}). Computation steps are detailed in \Cref{alg1} in Appendix~\ref{app4}.

\subsection{Automatic selection of acquisition data frame} \label{sec:autom_sel}
In clinical practice, multiple data acquisitions are often performed to image one tissue location.
Since these acquisitions may be subject to noise, such as that caused by hand/body motion affecting SoS reconstruction~\cite{schweizer2023robust}, selecting the best acquisition is crucial for diagnostic decisions and downstream tasks.
We propose to use the prediction uncertainty as a surrogate for such trust attribution to image reconstructions and hence acquisitions, cf.~\Cref{fig:pipeline_sos}c. 
Our goal is to automatically select one acquisition, and equivalently the corresponding reconstruction, out of N possibilities.

Given uncertainty estimations, one approach would be to select a reconstruction with minium uncertainty within an inclusion (given its operator annotated location from B-mode).
Frame \emph{selection }\emph{informed} by such \emph{inclusion} uncertainty is hereafter referred as SI$^\text{inc}_\text{MCD}$ or SI$^\text{inc}_\text{BVI}$, depending respectively on the uncertainty assessment approach.
Since absolute uncertainty value may vary significantly depending on the imaged tissue region, relying on the inclusion or total image uncertainty as a surrogate for assessing reconstruction confidence was found unreliable in our preliminary tests.
To address this, we introduce a normalized (\emph{relative}) uncertainty metric, defined as the absolute difference between the \emph{inclusion} uncertainty mean and the \emph{background} uncertainty mean, with background defined as the 5\,mm ring surrounding the inclusion. 
Frame selection using minimum \emph{relative} uncertainty is hereafter referred as SI$^\text{rel}_\text{MCD}$ or SI$^\text{rel}_\text{BVI}$, correspondingly. 
Computation of the above per-frame metrics and frame selection based on them are formalized in \Cref{alg2,alg3} in Appendix~\ref{app4}.

As baseline comparisons, we evaluate the diagnostic prediction performance \emph{without} informed frame selection.
For this, we use either simply the first acquisition (S1) or the third acquisition (S3) out of N. 
Additionally, we assess the results for choosing an acquisition at random (SR) for each patient.
We study these baselines for the reconstructions from the analytical method (S1$_\text{LBFGS}$, S3$_\text{LBFGS}$, SR$_\text{LBFGS}$), from the original VN (S1$_\text{VN}$, S3$_\text{VN}$, AR$_\text{VN}$) and from the two uncertainty estimation approaches (S1$_\text{MCD}$ \& S1$_\text{BVI}$, S3$_\text{MCD}$ \& S3$_\text{BVI}$, and SR$_\text{MCD}$ \& SR$_\text{BVI}$).

\section{Experimental setup}
\subsection{Simulated data}
\label{data_imp}

The VN to solve the inverse SoS problem requires supervision at training time. However, ground-truth local SoS distributions are unavailable and infeasible for in-vivo data, as this would require voxel-wise biomechanical testing of excised samples. Therefore, while testing is conducted on in-vivo patient data, during training we use only simulated data, as our utilized model-based reconstruction network enables us to bridge such large domain gap. For pulse-echo SoS reconstruction we use virtual source transmit (VS) sequences.
For each acquisition, 17 VS transmits are emitted, within 15 pairs of which the displacements are tracked~\cite{schweizer2023robust}.
Each simulated training sample, hence, contains 15 displacements maps and one ground truth SoS map of the imaged area.
For simulations and in-vivo, there is a same, single imaging operator, built by vertically stacking discretized acoustic path integral matrices~\cite{rau2021speed} for these said 15 pairs.
The data are simulated in two ways. 

First, we generate 10\,000 training samples using a ray-based (RB) approach, applying the known forward imaging model.  
To avoid inverse crime by overfitting to the imaging operator, we incorporate discretization and noise effects by following the data preprocessing in~\cite{vishnevskiy2019deep}, by actually simulating high-resolution noisy measurements \mbox{$d_\mathrm{hr}$ = $L_\mathrm{hr}x_\mathrm{hr}$ + $\epsilon$}, using a high-resolution imaging model $L_\mathrm{hr}$ and slowness map $x_\mathrm{hr}$, with $\epsilon$ Gaussian noise. 
SoS maps are then reconstructed on the original coarser grid, for the VN filters to learn to resolve noise and discretization artifacts.

Second, we generate 840 training samples using a sophisticated physics based simulation approach, based on the k-Wave Matlab toolbox. 
We simulated multi-static full-matrix single-element transmission, recording the RF echo signals at each receive (Rx) element for each Tx. 
We then retrospectively employ simulated Tx beamforming on such RF data to obtain the data for the desired 17 VS, for each and every training sample. 
The Rx beamforming is subsequently performed on these signals using delay-and-sum beamforming, followed by displacement tracking for each of the 15 pairs with a normalized cross-correlation algorithm. This process more accurately represents real data acquisition settings but is also more computationally costly.
For both cases, the ground truth SoS maps were designed to contain randomly shaped inclusions, created from deformations of random ellipses.
Half of these inclusions were filtered to generate smooth edges.
The background SoS value was varied slightly, to make the network more robust to such changes.
Additionally, 5\% of the samples did not include any inclusion to prevent the network from always predicting an inclusion. 
Finally, using both approaches a test set of 32 samples was generated to verify the \emph{reconstruction} accuracy of the \emph{uncertainty-aware models} against the baseline VN.
The data generation pipelines are illustrated in~\Cref{sim_data} in Appendix~\ref{app4}.
For more details on data generation, we refer the reader to~\cite{bernhardt2020training}. 

\subsection{Clinical data}
Clinical data was collected \emph{in vivo} from breast lesions in a study conducted at Kantonsspital Baden, Switzerland, with ethics approval and informed consent (EKNZ, Switzerland, BASEC 2020-01962). 
Each inclusion was imaged $N=3$ to $N=5$ times from different directions, with 17 VS transmits yielding 15 displacement maps as input to the VN for SoS reconstruction.  
We focus on differentiating FA and CA in patients in BI-RADS~4 class, which involves suspicious cases challenging for classification.
Out of 21 BI-RADS~4 patients, 8 had biopsy-confirmed CA and 13 had biopsy-confirmed FA.

\subsection{Implementation details}
The measurements and reconstructions are obtained on a  84\,x\,64 grid. 
As the analytical baseline, we employed the LBFGS algorithm. 
The models were implemented in Python and Tensorflow 1.11, and were ran on a 12\,GB NVIDIA TITAN\,Xp GPU. 
The training used mini-batch optimization, ADAM optimizer, 120\,000 iterations, and a batch size of 16 including a mix of 12 ray-based and 4 k-Wave samples. 
Dropout probability \textit{p} for MCD was set to 0.25. 
For BVI, kernel size $o=8$, $\alpha=0.1$, and $\beta=10$ were used.
For the initialization of the Gaussian parameters, we sampled the means normally with standard deviation of $10^{-2}$ and the covariance matrix diagonal uniformly between 0.9 and 1.1. 
Uncertainty-based frame selection comes at a computation cost, mainly because inference needs to be run for many sample repetitions, but also because Bayesian weights bring an overhead on the BVI model computation. 
We chose 100 samples, empirically based on a test image as a safe upper-bound to ensure convergence of uncertainty statistics. 
For constraints on computation time, a lower number of samples can be tested and may work as well.
\Cref{table_VNparam} in Appendix~\ref{app4} lists the parameter dimensions and the hyperparameter values used. 

\subsection{Evaluation}

Our selected frame is used for breast cancer differential diagnosis. 
For this diagnosis, we use the SoS contrast metric, which was shown to be effective in differentiating breast cancer.
This metric is defined here as $\Delta c = |c_\text{inc} - c_\text{bkg}|$, \ie the difference between the median inclusion SoS and the median background SoS, with the latter defined as the same 5\,mm ring as for relative uncertainty estimation.
For the \emph{in vivo} data, inclusion masks were annotated by an expert on B-mode images.

In simulations where the ground-truth SoS map $c$ is available, root-mean-square-error RMSE$=\sqrt{\frac{1}{n} \sum_{i=1}^{n} (c_i - \hat{c}_i)^2}$ over $n$ reconstructed pixels was calculated to assess reconstruction accuracy. 
For the clinical data, we assess performance in differentiating FA and CA cases, assuming a higher inclusion SoS contrast $\Delta c$ for the latter. 
Specifically, we analyze $\Delta c$ distributions between the two lesion groups. 
To quantify the classification performance, we compute Receiver Operating Characteristic (ROC) curves, Area Under the Curve (AUC) and F1 score, \ie $\frac{\mathrm{TP}}{\mathrm{TP}+\frac{1}{2}(\mathrm{FP}+\mathrm{FN})}$, where TP are true positives, FP false positives, TN true negatives, and FN false negatives.
We also report the sensitivity ($\frac{\mathrm{TP}}{\mathrm{TP}+\mathrm{FN}}$) and specificity ($\frac{\mathrm{TN}}{\mathrm{TN}+\mathrm{FP}}$), particularly at an operating point maximizing their sum.
For statistical significance, we used Wilcoxon rank-sum as an unpaired non-parametric test, since the data was observed not to be parametric given Levene's test for homogeneity of variance and the Shapiro test for normality.

\section{Results}
\subsection{Simulated data}

\begin{figure}
\centerline{\includegraphics[width=\columnwidth]{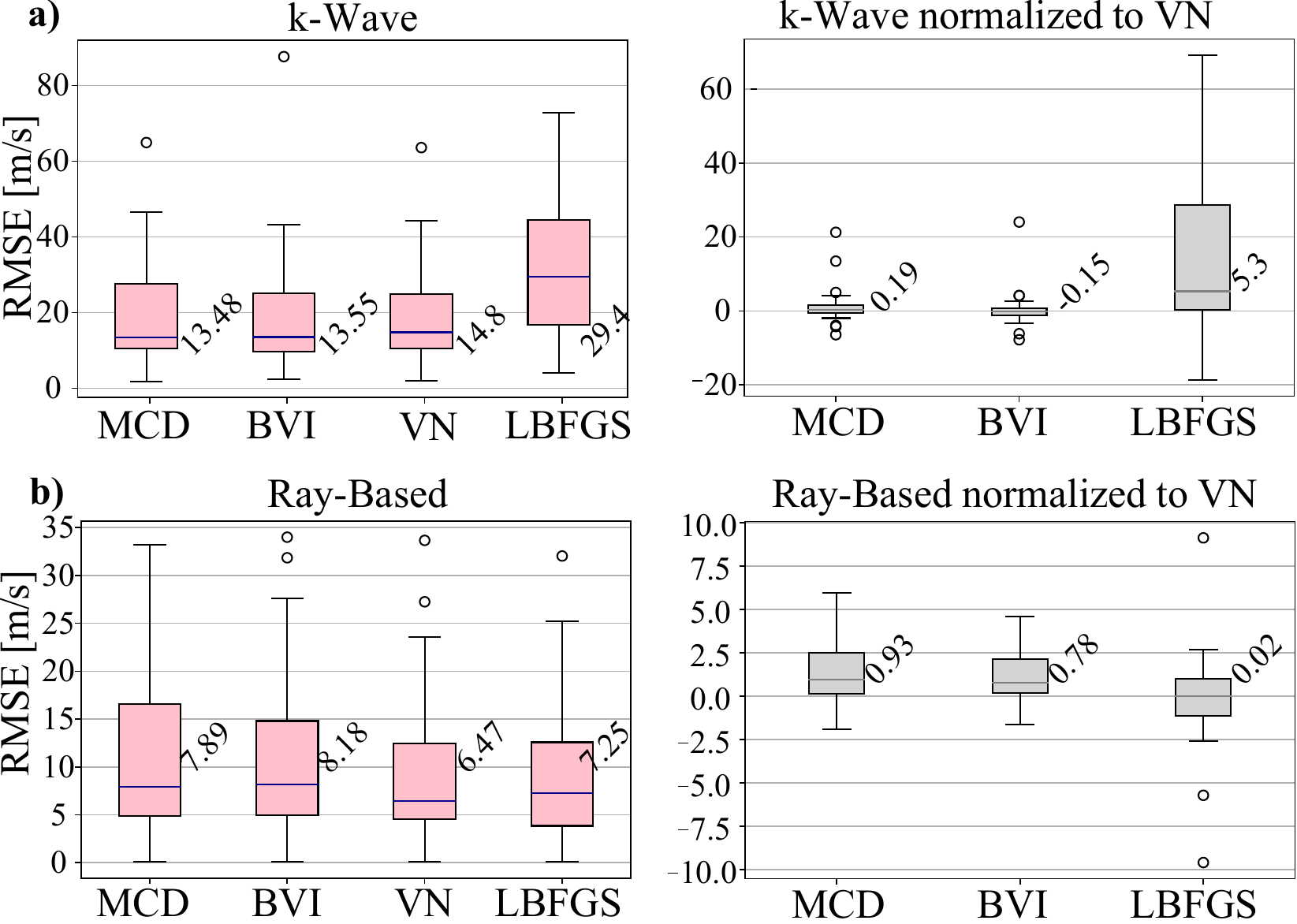}}
\caption{Reconstruction errors in a) the k-Wave test data, and b) the RB test data. (Left) RMSEs of the different reconstruction methods, and (right) image-wise differential RMSEs with respect to VN.}
\label{RMSE_tot}
\end{figure}

\begin{figure}
\centerline{\includegraphics[width=\columnwidth]{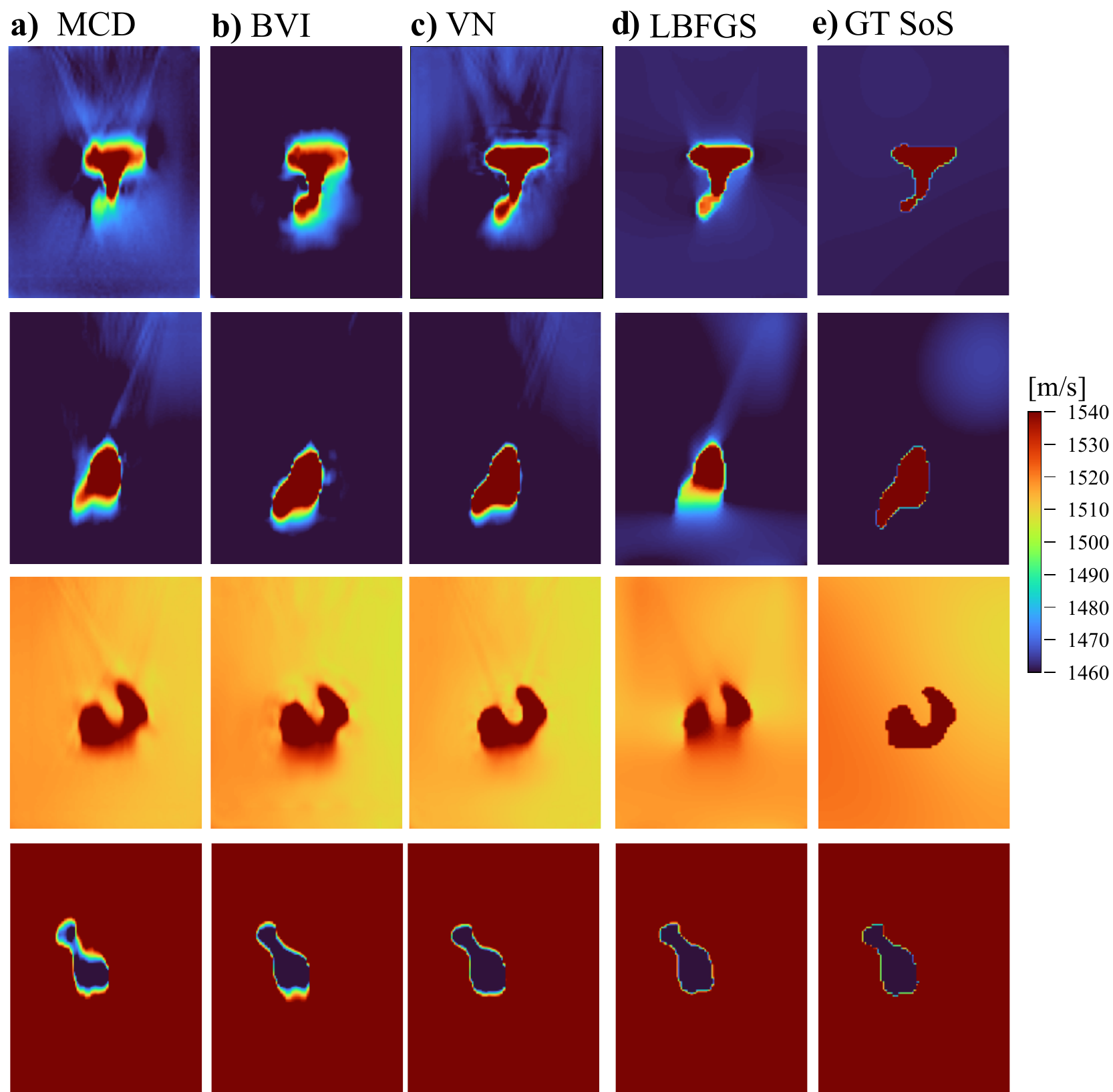}}
\caption{Examples reconstructions of four simulated ray-based numerical phantoms using each method (a-d), compared to the corresponding ground truth SoS maps (e).}
\label{tot_qual}
\end{figure}

We first tested whether adding uncertainty estimation on VN compromises its reconstruction quality.
For this, we used the simulation test set with ground truth maps, with the resulting RMSEs shown in \Cref{RMSE_tot}.
As seen, the reconstruction accuracies of VN augmented with the proposed uncertainty methods are not inferior to VN alone.
Reconstruction from k-Wave data is expectedly more challenging (as with higher RMSE) than RB data. 
Sample SoS reconstructions of RB data are shown in \Cref{tot_qual}.
\subsection{In vivo Data}
To use mean and standard deviation of uncertainty estimates, we first qualitatively checked the hypothesis of normal distribution of the network produced outputs.
\Cref{violin_dsos} in Appendix~\ref{app5} shows the flute plots of inclusion SoS contrast $\Delta c$ distribution across 100 samples for 8 sample patients, with the results showing normality.
For the frame selected based on uncertainty (based on pixel-wise sample SoS variation), $\Delta c$ is then calculated for the sample mean SoS image.
AUC and F1-score of the methods for differentiating CA from FA are shown in \Cref{combined_results}, 
\begin{table} 
\caption{Contrast-based lesion classification results of SoS reconstruction approaches based on selecting the first (S1), third (S3), and a random (SR) frame, as well as informed by inclusion uncertainty (SI$^\text{inc}$) and relative (inclusion normalized by background) uncertainty (SI$^\text{rel}$).}
\setlength{\tabcolsep}{7pt} 
\renewcommand{\arraystretch}{1.5} 
\centering 
\begin{tabular}{lc@{\ }c@{\ }c@{\ }cc@{\ }c@{\ }c@{\ }c} 
\toprule  & \multicolumn{4}{c}{\textbf{AUC (\%)}} & \multicolumn{4}{c}{\textbf{F1 Score (\%)}} \\
\toprule 
\textbf{Method} & \textbf{LBFGS} & \textbf{VN} & \textbf{BVI} & \textbf{MCD} & \textbf{LBFGS} & \textbf{VN} & \textbf{BVI} & \textbf{MCD} \\
\midrule 
\textbf{S1} & 54.81 & 49.04 & 54.81  & 58.65  & 55.17 & 61.54 & 58.82 & 57.14 \\
\midrule 
\textbf{S3}  & 59.62 & 54.81 & 51.90 & 51.90 & 63.59 & 51.94 & 51.82 & 51.82 \\
\midrule 
\textbf{SR}  & 63.47 & 45.21 & 54.42 & 59.61 & 64.02 & 59.28 & 58.81 & 63.59\\
\midrule 
\textbf{SI$^\text{inc}$} & -- & -- & 61.54 & 61.54 & --& --  & 63.16 & 61.54 \\
\midrule 
\textbf{SI$^\text{rel}$} & -- & -- & \textbf{{73.08}}  & \textbf{{75.96}} & --  & -- & \textbf{{80.00}} & \textbf{{70.59}} \\
\bottomrule 
\end{tabular} 
\label{combined_results}
\end{table}
indicating that that uncertainty-informed automatic frame selection shows potential for improving differential diagnosis performance compared to using a fixed (S1 or S3) or random (SR) acquisition.
These three uninformed selection approaches showing no systematic order of superiority across the four reconstruction methods (columns) shows that there is no major bias of classification results based on frame acquisition order.

\begin{figure*}
\centerline{\includegraphics[width=\textwidth]{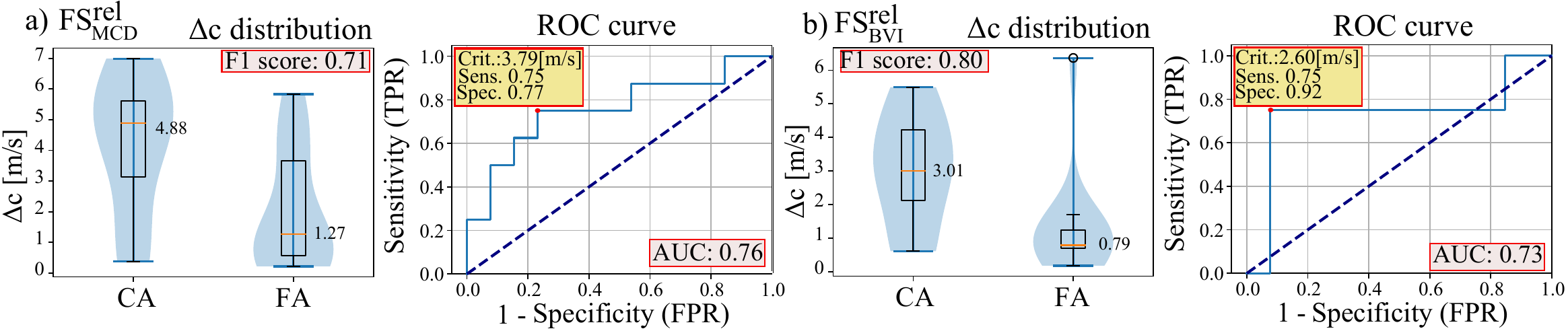}}
\caption{The $\Delta c$ distributions and ROC curves for CA vs. FA classification, for the uncertainty estimation approaches selecting the frames using  \textbf{(a)} SI$^\text{rel}_\text{MCD}$ for MCD-reconstructed frames and \textbf{(b)} SI$^\text{rel}_\text{BVI}$ for BVI-reconstructed frames.
The ROC curves display the classification criterion, sensitivity, and specificity at the point that maximizes the cumulative sensitivity and specificity.}
\label{ROC_DATA}
\end{figure*}

The $\Delta c$ distributions and ROC curves of the top two methods (SI$^\text{rel}_\text{MCD}$ and SI$^\text{rel}_\text{BVI}$) are presented in \Cref{ROC_DATA}, both of which are informed by relative uncertainty metric.
Both methods show an impressive improvement compared to the studied baselines, including the analytical method LBFGS and the baseline network architecture VN. 

\section{Discussion and conclusion}

Although uncertainty estimation has been studied widely in the literature, these methods are yet to find their clinical utility and niche application scenarios.
In this work, it is one of the first times we are proposing a use-case of uncertainty for automatic frame selection in scenarios where multiple repeated acquisitions are possible.
We have demonstrated this on VN based SoS reconstructions, for differential breast cancer diagnosis of patients in BI-RADS class 4, which involves suspicious cases that cannot be differentiated by other means such as the US B-mode. 
Accordingly, the differential diagnosis task involves classifying lesions that are visible and already identified by the radiologist on the US image.

We have adapted and trained VN for a specific VS Tx sequence used in the clinical data collection and have accordingly estimated SoS reconstruction uncertainty for the first time.
To that end, we used Monte Carlo and Bayesian techniques.
For the Bayesian approach, we have introduced a novel mathematical reformulation of the conventional loss function, which leads to a tractable implementation and increases training robustness.
We have shown that the proposed solutions are able to preserve (and even improves) the reconstruction accuracy of the baseline VN, while enabling an assessment of trust (confidence) to put on a particular reconstruction.
In this work, frame selection based on the proposed minimization of inclusion-to-background normalized uncertainty metric (SI$^\text{rel}$) proved most effective as a surrogate of trust in reconstructions. This may be due to such metric seeking for consistency in uncertainty between different image regions, i.e., inclusion and background.
Different problem settings may require different treatments, but our work highlights the challenge of directly comparing uncertainties across separate images, even of similar scenes.

Note that the VN framework learns the regularizers and priors, which makes it interpretable as opposed to blackbox learning algorithms.
Also, VN takes in the known imaging model matrix as input, so it does not have to learn it from data, making this method robust to train and easier to generalize across domain shift to in-vivo data, even from only simulated training.
Although the forward imaging model is fixed in VN, one could also learn and calibrate the imaging model in a restricted form using data, as shown in~\cite{bezek2023learning}.
Similarly to the conventional analytical methods (e.g., LBFGS), deep learned VN that we used also relies on the initial assumption of beamforming SoS.
This nevertheless can be estimated from displacement data~\cite{bezek_analytical_2023}, to be incorporated in the future for improved accuracy. 

In this manuscript, we propose a practical use-case of uncertainty estimation for selecting optimal frames from multiple data acquisitions.
We study this for breast cancer differential diagnosis based on VN-based US image reconstruction. 
Such solution can be extended and applied on many other problem settings and image modalities with pixel-wise regression-like outputs, where selecting one out of several possibilities is relevant.

\appendices

\section{Loss Definition and Exponential Weighting}
\label{app2}
To minimize gradients vanishing in the last layers (loops) during training, a weighted loss with exponential decrease of $\tau$ was used~\cite{vishnevskiy2019deep}:
\begin{align}
\mathcal{L}_{\text T} &= \sum_{k=1}^K \exp^{-\tau(K-k)}\left\|{x}_k-{x}^*\right\|_1 \notag \\
&\quad + \lambda_r \sum_{k=1}^K \sum_{j=1}^{N_\text{j}^{\phi_\text{k}}} 
\sqrt{\left(y_{j-1}^{\phi_k}-2 y_j^{\phi_k}+y_{j+1}^{\phi_k}\right)^2+\varepsilon}.
\end{align}
The second part of the loss corresponds to a smoothing of the transformation functions ${\phi_k}$, where ${N_\text{j}^{\phi_\text{k}}}$ is the number of knots for transformation function parametrization with the value $y_j^{\phi_k}$. 
Note that $\varepsilon$ is added for stability. 
The regularization weight $\lambda_{r}$ is used to balance smoothing and reconstruction losses. 
We refer the reader to~\cite{bernhardt2020training} for the detailed derivation of the data gradient and explanation of the network structure, which is out of the scope of this work. 

\section{Data Preprocessing and Implementation Details}
\label{app4}
\begin{figure}
\centerline{\includegraphics[width=\columnwidth]{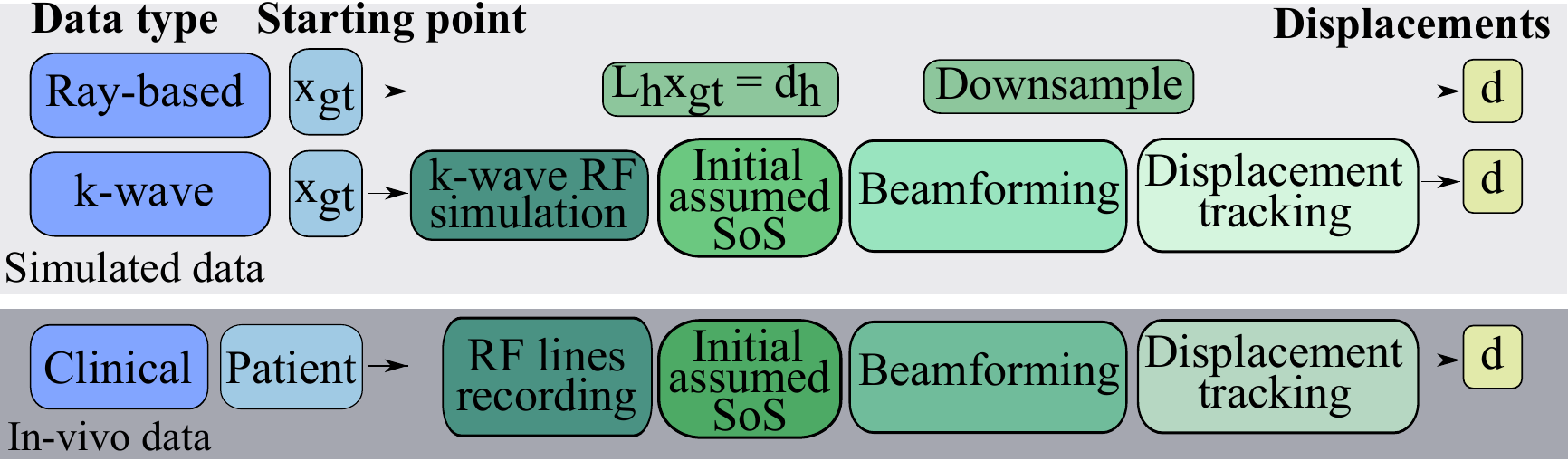}}
\caption{The displacement generation pipeline for the three datasets in this project. The first two simulation pipelines are for ray-based and k-wave data, where the ground truth is the starting point. The last row is for the \emph{in vivo} data, where the RF recordings are the starting point as the ground truth is unknown.}
\label{sim_data}
\end{figure}

\Cref{sim_data} provides a comparative overview of the steps followed for the data generation and preprocessing processes. 
The simulated data is obtained based on given ground truth SoS maps via two approaches as described in~\Cref{data_imp}. Overall, the final displacement d is the input to the VN and is acquired in the different ways shown below in both simulated and \emph{in vivo} data. 
\Cref{table_VNparam} includes a comprehensive list of the parameters used in the proposed methods.
\begin{table}
\centering
\caption{Parameters dimensions and hyperparameter values for the VN model and BVI \& MCD implementations.}\label{table_VNparam}
\begin{tabular}{lll}
\hline
{\textbf{Param.}} & \textbf{Description} & \textbf{Dimension} \\
\hline ${w}_k$ & Spatial weights & $57 \times 77 \times 32$ \\
${{r}_k}$ & Regularizers as conv. filters & $[8,8,32]$ \\
${{m}_k}$ & Dropout masks in MCD for conv. filters & $[8,8,32]$ \\
$\tilde{\mu}_{{r}_k}$ & Mean values of conv. filters & 32 \\
$\phi_{k}$ & Regularization term activation & 35 knots/layer \\
$\psi_{k}$ & Data term activation & 35 knots/layer\\
\hline
\textbf{Hyperpar.} & \textbf{Description} & \textbf{Value} \\
\hline
K & Number of layers & 20\\
$N_k$ & Number of filters & 32\\
 $\tau$ & Exponential weighting & 5\\
 $\lambda_{r}$ & Activation function smoothing & $10^5$\\
  $p$ & Dropout probability in MCD & 0.25\\
  $o$ & Kernel size in BVI & 8\\
  $\alpha$ & Prior for weight learning in BVI & 0.1\\
  $\beta$ & Kl regularization weight in BVI & 10\\
\hline
\end{tabular}
\end{table}

SoS and pixel-wise uncertainty estimation processes described in~\Cref{sec:unc} are formalized here in~\Cref{alg1}. 
Computations of the per-frame metrics from pixel-wise uncertainty estimates, \ie the minimum uncertinaty in the inclusion and the relative uncertainty of the inclusion normalized by the background, are described in~\Cref{alg2}. 
Frame selection based on these per-frame uncertainty surrogates is detailed in~\Cref{alg3}. A code snippet of the efficient algebraic reformulation introduced in~\Cref{log_det_eq} for the KL term of BVI loss in Tensorflow is as follows:
\lstset{language=Python, basicstyle=\ttfamily\small, breaklines=true}
\begin{lstlisting}
BVI_cost = (alpha_BVI * tf.linalg.trace(tf.matmul(st_dev_mat,
tf.transpose(st_dev_mat)))- 2 * tf.trace(tf.log(st_dev_mat))).
\end{lstlisting}

\begin{algorithm}
\caption{Uncertainty Estimation for SoS Reconstruction}
\label{alg1}
\begin{algorithmic}[1] \small

\Require Trained Variational Network (VN); input displacement data $d$; number of samples $K$.
\Ensure Reconstructed SoS image ($c$) and uncertainty.

\Procedure{Uncertainty Estimation } {Method}
    \If{Method = MCD}
        \State Set dropout probability $p$ during inference (same as during training).
        \For{$i = 1$ to $K$}
            \State Apply dropout mask to the VN.
            \State Compute $i$-th SoS reconstruction map: $c_i = \text{VN}(d)$.
        \EndFor
    \ElsIf{Method = BVI}
        \State VN with filters modeled as Gaussian distributions with mean $\mu$ and covariance $\Sigma$.
        \For{$i = 1$ to $K$}
            \State Sample weights $\theta_i \sim \mathcal{N}(\mu, \Sigma)$.
            \State Compute $i$-th SoS reconstruction map: $c_i = \text{VN}(d, \theta_i)$.
        \EndFor
    \EndIf
\EndProcedure

\Procedure{Compute Mean Pixel-Wise Reconstruction }{$c_i$}
    \For{each pixel $p = 1$ to $P$}
        \State $c_{p} = \frac{1}{K} \sum_{i=1}^{K} c_{i,p}$
    \EndFor
\EndProcedure

\Procedure{Compute Pixel-Wise Uncertainty } {$c_i, c_p$}
    \For{each pixel $p = 1$ to $P$}
        \State $\text{Uncertainty}_{p} = \sqrt{\frac{1}{K} \sum_{i=1}^{K} (c_{i,p} - c_{p})^2}$
    \EndFor
\EndProcedure

\end{algorithmic}
\end{algorithm}

\begin{algorithm}
\caption{Computing Per-Frame Metrics from Per-Pixel Uncertainty Values}
\label{alg2}
\begin{algorithmic}[1] \small

\Require Pixel-wise uncertainty estimates.
\Ensure Inclusion uncertainty ($u_\mathrm{inc}$) and relative uncertainty ($u_\mathrm{rel}$) metrics, computed per-frame as surrogates for trust attribution.

\Procedure{Uncertainty Metrics }{Uncertainty}
    \State Define lesion region $R_\mathrm{inc}$ based on a segmented mask.
    \State Define background region $R_\mathrm{bkg}$ as the 5 mm margin surrounding $R_\mathrm{inc}$.

    \State Compute mean uncertainty in $R_\mathrm{inc}$:  
    \[
    u_\mathrm{inc} = \frac{1}{|R_\mathrm{inc}|} \sum_{x \in R_\mathrm{inc}} \text{Uncertainty}(x)
    \]

    \State Compute mean uncertainty in $R_\mathrm{bkg}$:  
    \[
    u_\mathrm{bkg} = \frac{1}{|R_\mathrm{bkg}|} \sum_{x \in R_\mathrm{bkg}} \text{Uncertainty}(x)
    \]

    \State Compute relative uncertainty:  
    \[
    u_\mathrm{rel} = |u_\mathrm{inc} - u_\mathrm{bkg}|
    \]
\EndProcedure

\end{algorithmic}
\end{algorithm}

\begin{algorithm}
\caption{Frame Selection from Multiple Acquisitions}
\label{alg3}
\begin{algorithmic}[1] \small

\Require $N$: Total number of acquired frames; $u_{\mathrm{rel},i}$ and $u_{\mathrm{inc},i}$ for each frame $i$.
\Ensure Selected frames SI$^\mathrm{rel}$ (minimum relative uncertainty) and SI$^\mathrm{inc}$ (minimum inclusion uncertainty).

\Procedure{Select Frame with Minimum Inclusion Uncertainty }{$u_{inc}$}
    \State SI$^\mathrm{inc} = \arg\min\limits_{i \in \{1, .., N\}} u_{\mathrm{inc},i}$
\EndProcedure

\Procedure{Select Frame with Minimum Relative Uncertainty } {$u_{rel}$}
    \State SI$^\mathrm{rel} = \arg\min\limits_{i \in \{1, .., N\}} u_{\mathrm{rel},i}$
\EndProcedure

\end{algorithmic}
\end{algorithm}

\section{Probabilistic Inference Hypothesis in VN for SoS}
\label{app5}

\begin{figure}[!h]
\centerline{\includegraphics[width=\columnwidth]{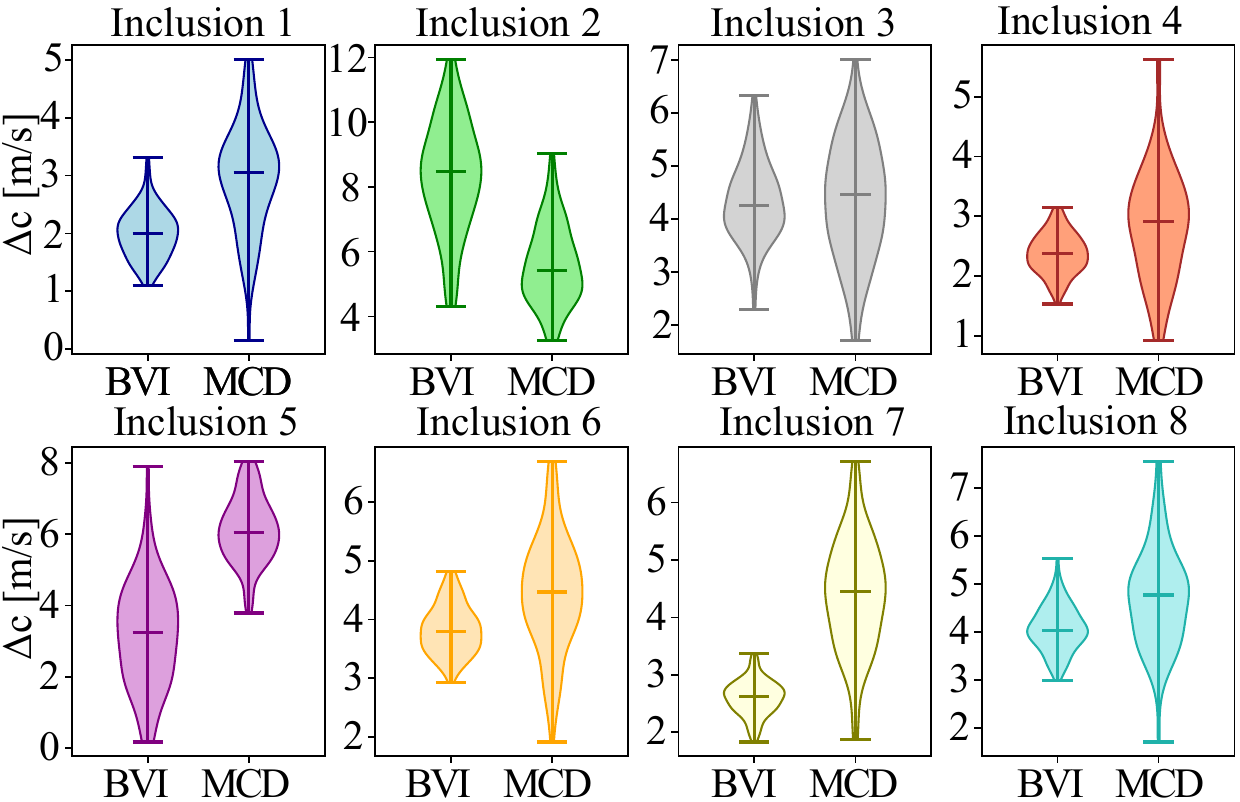}}
\caption{$\Delta c$ distributions in 100 samples of BVI and MCD in 8 clinical inclusions to show Gaussianity.}
\label{violin_dsos}
\end{figure}

The studied uncertainty estimation methods rely on multiple samples of a posterior distribution of the model. \Cref{violin_dsos} shows an example of the $\Delta c$ distributions acquired from each method on 8 different inclusions of the clinical data. From the Gaussianity of the distribution, we can conclude that the mean and the standard deviation are reasonable estimates for the SoS reconstruction and uncertainty computation. 

\bibliography{arxiv}

\end{document}